\documentclass{article}

\usepackage{nips06,times}
\usepackage{graphicx}
\usepackage{amssymb,amsmath} 
\usepackage{subfigure} 

\newcommand{\eqnref}[1]{Eq.~(\ref{#1})}

\newcommand{\figref}[1]{Fig.~\ref{#1}}

\newcommand{\hmu}	{h_\mu} 
\newcommand{\hmuL}	{h_{\mu L}} 
\newcommand{\hmuk}	{h_{\mu k}} 
\newcommand{\KLd}	{\mathcal{D}} 
\newcommand{\Asize}[0]{\vert \mathcal{A} \vert} 
\newcommand{\hk}[0]{ \overleftarrow{s}^k} 
\newcommand{\htl}[2]{ \overleftarrow{s}_{#1}^{#2}} 
\newcommand{\nsk}[0]{n(\overleftarrow{s}^k)} 
\newcommand{\nsks}[0]{n(\overleftarrow{s}^k s)} 
\newcommand{\ask}[0]{\alpha(\overleftarrow{s}^k)} 
\newcommand{\asks}[0]{\alpha(\overleftarrow{s}^k s)} 
\newcommand{\MPk}[0]{\mathbf{\theta}_k} 
\newcommand{\MCk}[0]{\mathbf{M}_k} 
\newcommand{\avg}[2]{\mathbf{E}_{\textrm{#2}}[#1]} 
\newcommand{\psk}[0]{p(\overleftarrow{s}^k)}  
\newcommand{\psks}[0]{p(s\vert \overleftarrow{s}^k)}  
\newcommand{\qsk}[0]{q(\overleftarrow{s}^k)}  
\newcommand{\qsks}[0]{q(s\vert \overleftarrow{s}^k)}  

\newcommand{\pme}[0]{PME} 

\title{How Random is a Coin Toss?\\
Bayesian Inference and\\
the Symbolic Dynamics of Deterministic Chaos}

\author{
Christopher C. Strelioff \\
Center for Complex Systems Research and Department of Physics\\
University of Illinois at Urbana-Champaign\\
Urbana, IL 61801 \\
and\\
Center for Computational Science and Engineering \\
University of California at Davis \\
Davis, CA 95616 \\
\texttt{streliof@uiuc.edu} \\
\AND
James P. Crutchfield\\
Center for Computational Science and Engineering and Physics Department \\
University of California at Davis \\
Davis, CA 95616 \\
\texttt{chaos@cse.ucdavis.edu}
}

\begin{document}

\maketitle

\begin{abstract}
Symbolic dynamics has proven to be an invaluable tool in analyzing the
mechanisms that lead to unpredictability and random behavior in nonlinear
dynamical systems. Surprisingly, a discrete partition of continuous
state space can produce a coarse-grained description of the behavior
that accurately describes the invariant properties of an underlying
chaotic attractor.  In particular, measures of the rate of information
production---the topological and metric entropy rates---can be estimated
from the outputs of Markov or generating partitions. Here we develop
Bayesian inference for $k$-th order Markov chains as a method to finding
generating partitions and estimating entropy rates from finite samples
of discretized data produced by coarse-grained dynamical systems.
\end{abstract}

\section{Introduction}

Research on chaotic dynamical systems during the last forty years produced a
new vision of the origins of randomness.  It is now widely understood that
observed randomness can be generated by low-dimensional deterministic systems
that exhibit a chaotic attractor. Today, when confronted with what appears
to be a high-dimensional stochastic process, one now asks whether or not the
process is instead a hidden low-dimensional, but nonlinear dynamical system.
This awareness, though, requires a new way of looking at apparently random
data since chaotic dynamics are very sensitive to the measurement
process~\cite{Bollt2000}, which is both a blessing and a curse, as
it turns out.

Symbolic dynamics, as one of a suite of tools in dynamical systems theory, in
its most basic form addresses this issue by considering a coarse-grained view
of a continuous dynamics.~\footnote{For a recent overview
consult~\cite{BLHao1998} and for a review of current applications
see~\cite{Daw2002} and references therein.}  In this sense, any
finite-precision instrument that measures a chaotic system induces a
symbolic representation of the underlying continuous-valued behavior.

To effectively model time series of discrete data from a continuous-state
system two concerns must be addressed.  First, we must consider the measurement
instrument and the representation of the true dynamics which it provides.
Second, we must consider the inference of models based on this data.  The
relation between these steps is more subtle than one might expect. As we
will demonstrate, on the one hand, in the measurement of chaotic data, the
instrument should be designed to maximize the entropy rate of the resulting
data stream. This allows one to extract as much information from each
measurement as possible. On the other hand, model inference strives to
minimize the apparent randomness (entropy rate) over a class of alternative
models. This reflects a search for determinism and structure in the data.

Here we address the interplay between optimal instruments and optimal models
by analyzing a relatively simple nonlinear system.  We consider the design
of binary-output instruments for chaotic maps with additive noise. We then
use Bayesian inference of a $k$-th order Markov chain to model the resulting
data stream. Our model system is a one-dimensional chaotic map with additive
noise~\cite{Crutchfield1982,Crutchfield1983}
\begin{equation}
	x_{t+1} = f(x_{t}) + \xi_{t} ~,
\end{equation}
where $t = 0, 1, 2, \ldots$, $x_t \in [0,1]$, and
$\xi_{t} \sim \textrm{N}(0,\sigma^2)$ is Gaussian random variable with mean
zero and variance $\sigma^2$. To start we consider the design of instruments
in the zero-noise limit. This is the regime of most previous work in symbolic
dynamics and provides a convenient frame of reference.

The construction of a symbolic dynamics representation of a continuous-state
system goes as follows \cite{BLHao1998}. We assume time is discrete and
consider a map $f$ from the \emph{state space} $M$ to itself
$f:M \rightarrow M$. This space can partitioned into a finite set
$\mathcal{P}=\{I_i : \cup_i I_i = M , I_i \cap I_j = 0 , i \neq j\}$ of
nonoverlapping regions in many ways. The most powerful is called a
\emph{Markov partition} and must satisfy two conditions.  First, the image of
each region $I_i$ must be a union of intervals:
$f(I_i) = \cup_j \, I_j , \forall \, i$.  Second, the map $f(I_i)$, restricted
to an interval, must be one-to-one and onto.  If a Markov
partition cannot be found for the system under consideration, the
next best coarse-graining is called a \emph{generating
partition}.  For one dimensional maps, these are often easily
found using the extrema of $f(x)$---its \emph{critical points}.
The critical points in the map are used to divide the state space
into intervals $I_i$ over which $f$ is monotone.
Note that Markov partitions are generating, but the converse is
not generally true.

Given any partition $\mathcal{P}= \{I_i\}$, then, a series of continuous-valued
states $\mathbf{X}=x_{0}x_{1} \ldots x_{N-1}$ can be projected onto its symbolic
representation $\mathbf{S}=s_{0}s_{1} \ldots s_{N-1}$. The latter is simply the
associated sequence of partition-element indices. This is done by defining
an operator $\pi(x_t)=s_t$ that returns a unique symbol $s_t = i$ for each
$I_i$ from an alphabet $\mathcal{A}$ when $x_t \in I_i$.

The central result in symbolic dynamics establishes that, using a generating
partition, increasingly long sequences of observed symbols identify smaller
and smaller regions of the state space. Starting the system in such a region
produces the associated measurement symbol sequence. In the limit of infinite
symbol sequences, the result is a discrete-symbol representation of a
continuous-state system---a representation that, as we will show, is often
much easier to analyze. In this way a chosen partition creates a symbol sequence
$\pi(\mathbf{X}) = \mathbf{S}$ which describes the continuous dynamics as a
sequence of symbols. The choice of partition then is equivalent to our
instrument-design problem.

The effectiveness of a partition (in the zero noise limit) can be quantified
by estimating the entropy rate of the resulting symbolic sequence. To do this
we consider length-$L$ \emph{words}
$\mathbf{s}^L = s_{i}s_{i+1} \ldots s_{i+L-1}$.
The \emph{block entropy} of length-$L$ sequences obtained from partition
$\mathcal{P}$ is then
\begin{equation}
  H_{L}(\mathcal{P}) =
  - \sum_{\mathbf{s}^L \in \mathcal{A}^L} \, p(\mathbf{s}^L)
  \log_2 p(\mathbf{s}^L) ~,
\label{eqn:block_L_entropy}
\end{equation}
where $p(\mathbf{s}^L)$ is the probability of observing the word
$\mathbf{s}^L \in \mathcal{A}^L$. From the block entropy the
\emph{entropy rate} can be estimated as the following limit
\begin{equation}
\hmu(\mathcal{P}) = \lim_{L \to \infty} \frac{ H_{L}(\mathcal{P}) }{L} ~.
\label{eqn:entropy_rate_partition}		
\end{equation}
In practice it is often more accurate to calculate the length-$L$ estimate
of the entropy rate using
\begin{equation}
\hmuL (\mathcal{P}) = H_L (\mathcal{P})  - H_{L-1} (\mathcal{P}) ~.
\label{eqn:HmuLengthL}
\end{equation}

Another key result in symbolic dynamics says that the entropy of
the original continuous system is found using generating
partitions \cite{Kolmogorov1958,Kolmogorov1959}. In particular,
the true entropy rate maximizes the estimated entropy rates:
\begin{equation}
  h_{\mu} = \underset{ \{\mathcal{P}\} }{\text{max}} \;
  h_{\mu}(\mathcal{P}) ~.
\label{eqn:kolmogorov_entropy_rate}
\end{equation}
Thus, translated into a statement about experiment design, the results tell
us to design an instrument so that it maximizes the observed entropy rate.
This reflects the fact that we want each measurement to produce the
most information possible.

As a useful benchmark on this, useful only in the case when we know $f(x)$,
\emph{Piesin's Identity}~\cite{Piesin1977} tells us that the value of
$h_{\mu}$ is equal the sum of the positive Lyapunov characteristic exponents:
$h_{\mu} = \sum_{i} \lambda_i^{+}$. For one-dimensional maps there is a
single Lyapunov exponent which is numerically estimated from the map $f$
and observed trajectory $\{ x_t \}$ using
\begin{equation}
  \lambda = \lim_{N \to \infty} \frac{1}{N}
  \sum_{t=1}^{N} \log_2 \vert f^\prime (x_t)\vert ~.
\label{eqn:lyapunov_exponent}
\end{equation}

Taken altogether, these results tell us how to design our instrument for
effective observation of deterministic chaos. Notably, in the presence of
noise no such theorems exist. However, \cite{Crutchfield1982,Crutchfield1983}
demonstrated the methods developed above are robust in the presence of noise.

In any case, we view the output of the instrument as a stochastic process. A
sample realization $D$ of length $N$ with measurements taken from a finite
alphabet is the basis for our inference problem:
$D = s_0 s_1 \ldots s_{N-1} \; , \; s_t \in \mathcal{A}$.
For our purposes here, the sample is generated by a partition of
continuous-state sequences from iterations of a one-dimensional
map and that states are on a chaotic attractor. This means, in
particular, that the stochastic process is stationary.
We assume, in addition, that the alphabet is binary $\mathcal{A} = \{ 0,1 \}$.

\section{Bayesian inference of $k$-th order Markov chains}

Given a method for instrument design the next step is to estimate a model from
the observed measurements. Here we choose to use the model class of $k$-order
Markov chains and Bayesian inference as the model estimation and selection
paradigm.

The $k$-th order Markov chain model class makes two strong assumptions about
the data sample. The first is an assumption of finite memory.  In other words,
the probability of $s_t$ depends only on the previous $k$ symbols in the data
sample. We introduce the more compact notation
$\htl{t}{k} = s_{t-k-1} \ldots s_t$ to indicate a length-$k$
sequence of measurements ending at time $t$. The finite memory assumption is
then equivalent to saying the probability of the observed data can be factored
into the product of terms with the form $p(s_t \vert \htl{t}{k})$.  The second
assumption is stationarity. This means the probability of observed sequences
does not change with the time position in the data sample:
$p(s_t \vert \htl{t}{k}) = \psks$ for any index $t$. As noted above, this
assumption is satisfied by the data streams produced.  The first assumption,
however, is often not true of chaotic systems. They can generate time series
with infinitely long temporal correlations. Thus, in some cases, we may be
confronted with out-of-class modeling.

The $k$-th order Markov chain model class $\MCk$ has a set of parameters
$\MPk = \{ \psks : s \in \mathcal{A}, \overleftarrow{s}^k \in \mathcal{A}^k \}$.
In the Bayesian inference of the model parameters $\MPk$ we must write down
the likelihood $P(D\vert \MPk, \MCk)$ and the prior $P(\MPk \vert \MCk)$ and
then calculate the evidence $P(D\vert \MCk)$. The posterior distribution
$P(\MPk \vert D, \MCk)$ is obtained from Bayes' theorem
\begin{equation}
\label{eqn:bayes}
  P\left( \MPk \vert D, \MCk \right) = \frac{ P\left( D \vert \MPk , \MCk \right) \; 
  P\left( \MPk \vert \MCk \right) }{ P\left( D \vert \MCk \right) } ~.
\end{equation}
The posterior describes the distribution of model parameters $\MPk$ given
the model class $\MCk$ and observed data $D$. From this the expectation of
the model parameters can be found along with estimates of the uncertainty
in the expectations. In the following sections we outline the specification
of these quantities following~\cite{Baldi2001,MacKay2003}.

\subsection{Likelihood}

Within the $\MCk$ model class, the likelihood of an observed data sample is
given by
\begin{equation}
\label{eqn:likelihood}
  P(D\vert \MPk, \MCk) = \prod_{ s \in \mathcal{A} } \prod_{ \hk \in \mathcal{A}^{k} } 
	p( s \vert \hk )^{\nsks} ~,
\end{equation}
where $\nsks$ is the number of times the word $\hk s$ occurs in sample $D$.
We note that~\eqnref{eqn:likelihood} is conditioned on the start sequence
$\hk_k = s_0s_1\ldots s_{k-1}$.

\subsection{Prior}

The prior is used to describe knowledge about the model class. In the case
of the $\MCk$ model class, we choose a product of Dirichlet
distributions---the so-called \emph{conjugate prior} \textbf{citations}.
Its form is
\begin{equation}
  P(\MPk \vert \MCk ) = \prod_{\hk \in \mathcal{A}^{k}} \
    \frac{ \Gamma( \ask  ) }{ \prod_{s\in\mathcal{A}} \Gamma( \asks ) }
    \, \delta \mathbf{(}1-\sum_{s\in\mathcal{A}} p( s \vert \hk ) \mathbf{)} 
	\, \prod_{s\in\mathcal{A}} p( s \vert \hk )^{\asks-1} ~,
\label{eqn:prior}
\end{equation}
where $\ask = \sum_{s \in \mathcal{A}} \asks$ and $\Gamma(x)$ is the gamma
function. The prior's parameters
$\{ \asks : s\in \mathcal{A}, \hk \in \mathcal{A}^k \}$ are assigned to
reflect knowledge of the system at hand and must be real and positive.
An intuition for the meaning of the parameters can be obtained by considering
the mean of the Dirichlet prior, which is
\begin{equation}
	\avg{\psks}{prior}  =  \frac{ \asks }{ \ask } ~.
\label{eqn:prior_mean} 
\end{equation}
In practice, a common assignment is $\asks = 1$ for all parameters.  This
produces a uniform prior over the model parameters, reflected by the
expectation $\avg{p(s\vert \hk)}{prior} = 1/ \vert \mathcal{A} \vert $.
Unless otherwise stated, all inference in the following uses the uniform prior.

\subsection{Evidence}

The evidence can be seen as a simple normalization term in Bayes theorem.
However, when model comparison of different orders and estimation of entropy
rates is considered, this term becomes a fundamental part of the analysis.
The evidence is defined
\begin{equation}
P(D\vert \MCk ) = \int \; d\MPk \; P(D\vert \MPk, \MCk) P(\MPk \vert \MCk ) ~,
\label{eqn:evidence_defn}
\end{equation}
It gives the probability of the data $D$ given the model order $\MCk$. For
the likelihood and prior derived above, the evidence is found analytically
\begin{equation}
P(D\vert \MCk) = \prod_{\hk \in \mathcal{A}^{k}} \; 
  \frac{ \Gamma(\ask) }{ \prod_{s\in \mathcal{A}} \Gamma(\asks)}
  \frac{ \prod_{s\in \mathcal{A}} \Gamma(\nsks+\asks) }{ \Gamma(\nsk+\ask) } ~.
\label{eqn:evidence}
\end{equation}

\subsection{Posterior}

The posterior distribution is constructed from the elements derived above according to Bayes' theorem~\eqnref{eqn:bayes}, resulting in a product of Dirichlet distributions.  This form is a result of choosing the conjugate prior and generates the familiar form
\begin{eqnarray}
P(\MPk\vert D, \MCk) & = & 	\prod_{\hk \in \mathcal{A}^{k}}
\frac{ \Gamma( \nsk + \ask  ) }{ \prod_{s\in\mathcal{A}} \Gamma( \nsks + \asks ) } 
\nonumber \\
 & \times &		
   \delta \mathbf{(}1-\sum_{s\in\mathcal{A}} p( s \vert \hk )\mathbf{)}
  \prod_{s\in\mathcal{A}} p( s \vert \hk )^{\nsks + \asks - 1} ~.
\label{eqn:posterior}
\end{eqnarray}
The mean for the model parameters $\MPk$ according to the posterior
distribution is then
\begin{equation}
	\avg{\psks}{post} =  \frac{ \nsks + \asks }{ \nsk + \ask } ~.
\label{eqn:posterior_mean} 
\end{equation}

Given these estimates of the model parameters $\MPk$, the next step is
decide which order $k$ is best for a given data sample.

\section{Model comparison of orders $k$}

Bayesian model comparison is very similar to the parameter estimation process
discussed above.  We start by enumerating the set of model orders to consider
$\mathcal{M} = \{ \MCk : k \in [k_{min}, k_{max}]\}$.  The probability of a
particular order can be found by considering two factorings of the joint
distribution $P(\MCk, D \vert \mathcal{M})$. Solving for the probability of
a particular order we obtain
\begin{equation}
  P(M_{k} \vert D , \mathcal{M} )
    = \frac{ P(D \vert M_{k}, \mathcal{M} ) P(M_{k} \vert \mathcal{M} ) }
	{P(D \vert \mathcal{M})} ~.	
\label{eqn:model_comparison}
\end{equation}   
where the denominator is given by the sum
$P(D \vert \mathcal{M}) = \sum_{M_{k'} \in \mathcal{M}} P(D \vert M_{k'}, \mathcal{M} )P(M_{k'} \vert \mathcal{M} )$.
This expression is driven by two components: the evidence
$P(D \vert M_{k}, \mathcal{M} )$ derived above and the prior over model
orders $P(M_{k} \vert \mathcal{M} )$.  Two common priors are a uniform prior
over orders and an exponential penalty for the size of the model
$P(M_{k} \vert \mathcal{M} ) = \exp(- \vert \MCk \vert)$.  For a $k$-th
order Markov chain the size of the model, or number of free parameters,
is given by
$\vert M_k \vert = \vert \mathcal{A} \vert^k(\vert \mathcal{A} \vert-1)$.
To illustrate the method we will consider only the prior over orders $k$
with a penalty for model size.

\section{Estimating Entropy Rates}

The entropy rate of an inferred Markov chain can be estimated by
extending the method for independent identically
distributed (IID) models of discrete data~\cite{Samengo2002}
using \emph{type theory}~\cite{Cover1991}. In simple terms, type theory shows
that the probability of an observed sequence can be suggestively rewritten in
terms of the \emph{Kullback-Leibler} (KL) distance and the entropy rate
\eqnref{eqn:entropy_rate_partition}. This form suggests a connection to statistical mechanics
and this, in turn, allows us to find average information-theoretic quantities
over the posterior by taking derivatives. In the large data limit, the KL
distance vanishes and we are left with the desired estimation of the Markov
chain's entropy rate. The complete development is beyond our scope here, but
will appear elsewhere. However, we will provide a brief sketch of the
derivation and quote the resulting estimator.

The connection we draw between inference and information theory starts by
considering the product of the prior~\eqnref{eqn:prior} and
likelihood~\eqnref{eqn:likelihood}
$P(\MPk\vert \MCk)P( D\vert \MPk, \MCk)=P( D, \MPk\vert \MCk)$.
This product forms a joint distribution over the observed data $D$ and model
parameters $\MPk$ given the model class $\MCk$.  Writing the normalization
constant from the prior as $Z$ to save space, this joint distribution can
be written, without approximation, in terms of conditional relative entropies
$\KLd[ \cdot \| \cdot ]$ and entropy rates $\hmu[\cdot]$
\begin{equation}
  P( D, \MPk\vert \MCk)
  =  Z \, 2^{-\beta_k \mathbf{(} \KLd [Q \| P ] + \hmu [Q]\mathbf{)}}
   2^{+\Asize^{k+1} \mathbf{(} \KLd [ U \| P ] + \hmu [U]\mathbf{)}} ~,
\label{eqn:info_prior_likelihood}
\end{equation}
where $\beta_k = \sum_{\hk,s} \nsks + \asks$. The set of probabilities used
above are
\begin{eqnarray}
  Q & =& \{ \qsk = \frac{\nsk+\ask}{\beta_k},
  \qsks = \frac{\nsks + \asks}{\nsk + \ask} \}
\label{eqn:pme_estimates} \\
  U & =& \{ \qsk = \frac{1}{\Asize}, \qsks = \frac{1}{\Asize} \} ~,
\label{eqn:uniform_estimates} 
\end{eqnarray}
where $Q$ is the distribution defined by the posterior mean, $U$ is a
uniform distribution, and $P = \{ \psk, \psks \}$ are the ``true''
parameters given the model class. The information theory quantities are
given by 
\begin{eqnarray}
\KLd [ Q \| P ] & = & \sum_{s, \hk} \qsk \qsks \log_2 \frac{\qsks}{\psks}
\label{eqn:conditional_KL_div} \\
\hmu [ Q ] 	& = & - \sum_{s, \hk} \qsk \qsks \log_2 \qsks ~. 
\label{eqn:entropy_rate}
\end{eqnarray}

The form of~\eqnref{eqn:info_prior_likelihood} and its relation to the evidence
motivates the connection to statistical mechanics. If we think of the evidence
$P(D \vert \MCk) = \int d\MPk P( D, \MPk\vert \MCk)$, as a
\emph{partition function} $\mathcal{Z} = P( D \vert \MCk)$, the
\emph{free energy} for the inference problem is simply
$\mathcal{F} = - \log \mathcal{Z}$.
Using conventional techniques from statistical mechanics, the expectation and
variance of $\KLd [Q \| P ] + \hmu [Q]$ are obtained by taking derivatives of
$\mathcal{F}$ with respect to $\beta_k$. In this sense
$\KLd [Q \| P ] + \hmu [Q]$ plays the role of an internal energy and
$\beta_k$ is comparable to an inverse temperature. We take advantage of the
known form for the evidence provided in~\eqnref{eqn:evidence} to calculate
the desired expectation resulting in 
\begin{eqnarray}
\avg{\, D[Q \| P ] + h[Q] \, }{post}  & = & \frac{1}{\log 2} 
	\sum_{\hk} \qsk \psi^{(0)} \left[ \beta_k \qsk \right]
	\label{eqn:average_info} \\
	& - & \frac{1}{\log 2} \sum_{\hk,s} \qsk \qsks \psi^{(0)}
	\left[ \beta_k \qsk \qsks \right] ~,\nonumber
\end{eqnarray} 
where the polygamma function is defined as
$\psi^{(n)}(x) = d^{n+1}/dx^{n+1} \log \Gamma(x)$.
The meaning of the terms on the RHS of~\eqnref{eqn:average_info} is not
immediately clear. However, we can use an expansion of the $n=0$ polygamma
function $\psi^{(0)}(x) = \log x - 1/2x + \mathcal{O}(x^{-2})$, which is
valid for $x \gg 1$, to find the asymptotic form
\begin{equation}
\avg{\, \KLd [Q \| P ] + \hmu [Q] \, }{post} = H_{k+1}[Q] - H_{k}[Q]
  + \frac{1}{2\beta_k} \Asize^k(\Asize -1 ) ~.
\label{eqn:average_info_asymptotic}
\end{equation}
From this expansion we can see that the first two terms make up the entropy
rate $\hmuk[Q] = H_{k+1}[Q] - H_{k}[Q]$. And the last term must be
associated with the conditional relative entropy between the posterior mean
estimate (\pme) distribution $Q$ and the true distribution $P$.

\section{Experimental Setup}

Now that we have our instrument design and model inference methods fully
specified we can describe the experimental setup used to test them. Data
from simulations of the logistic one-dimensional map, given by
$f(x_t) = r x_t(1-x_t)$,
at the chaotic value of $r=4.0$ was the basis for the analysis. A fluctuation
level of $\sigma = 10^{-3}$ was used for the added noise.  A random initial
condition in the unit interval was generated and one thousand transient steps,
not analyzed, were generated to find a typical state on the chaotic attractor.
Next, a single time series $x_0, x_1, \ldots, x_{N-1}$ of length $N=10^4$ was
produced.

A family of binary partitions
$\mathcal{P} (d) = \{ ``0'' \sim x \in [0,d) , ~ ``1'' \sim x \in [d,1] \}$
of the continuous-valued states was produced for two hundred decision points
$d$ between $0$ and $1$. That is, values in the state time series which
satisfied $x_t < d$ were assigned symbol $0$ and all others were assigned $1$.
Given the symbolic representation of the data for a particular partition
$\mathcal{P}(d)$, Markov chains from order $k=1$ to $k=8$ were inferred and
model comparison was used to select the order that most effectively described
the data. Then, using the selected model, values of entropy rate $\hmu(d)$
versus decision point $d$ were produced. 

\section{Results}

\begin{figure}[ht]
\centering
\subfigure[Instrument design.] 
{
    \label{fig:sub:a}
    \includegraphics[width=0.47\textwidth]{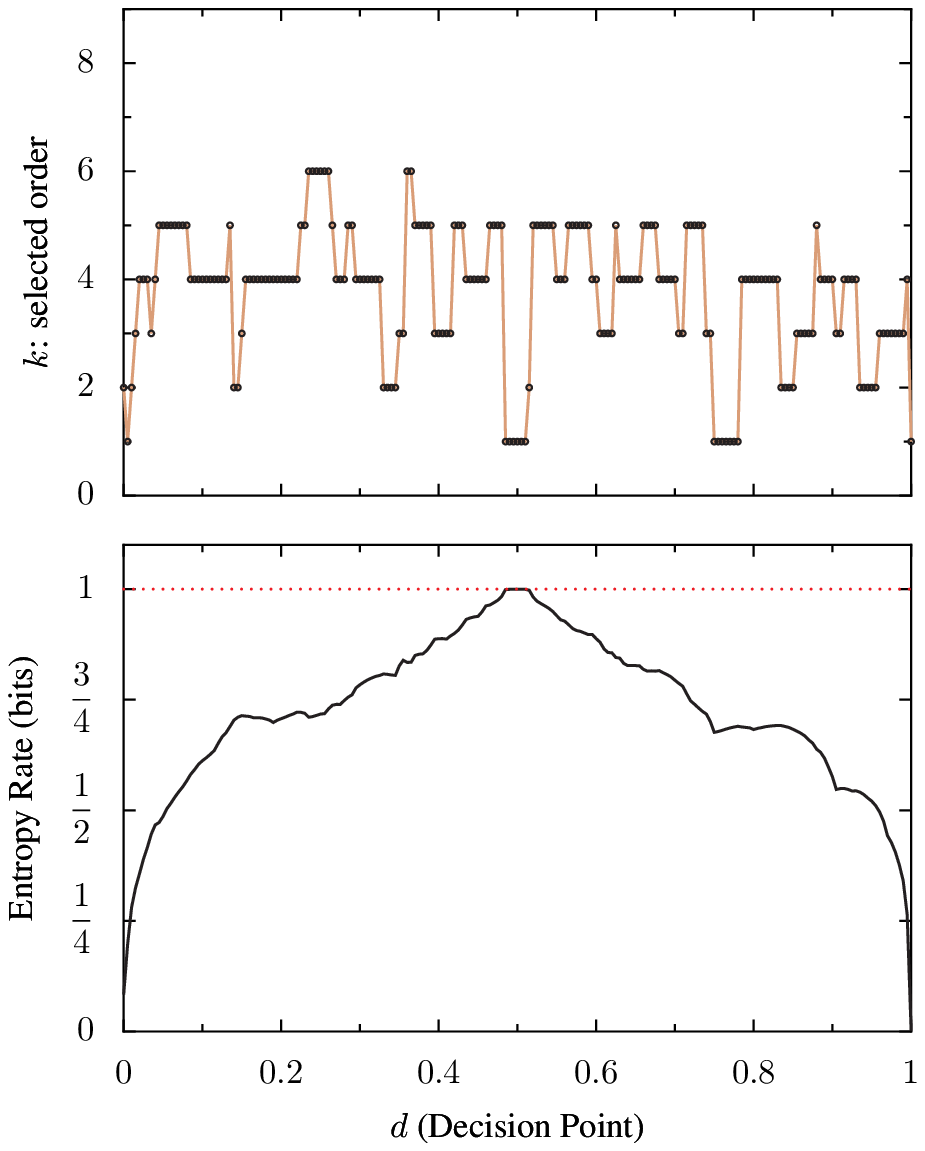}
}
\hfill
\subfigure[Model selection.] 
{
    \label{fig:sub:b}
    \includegraphics[width=0.49\textwidth]{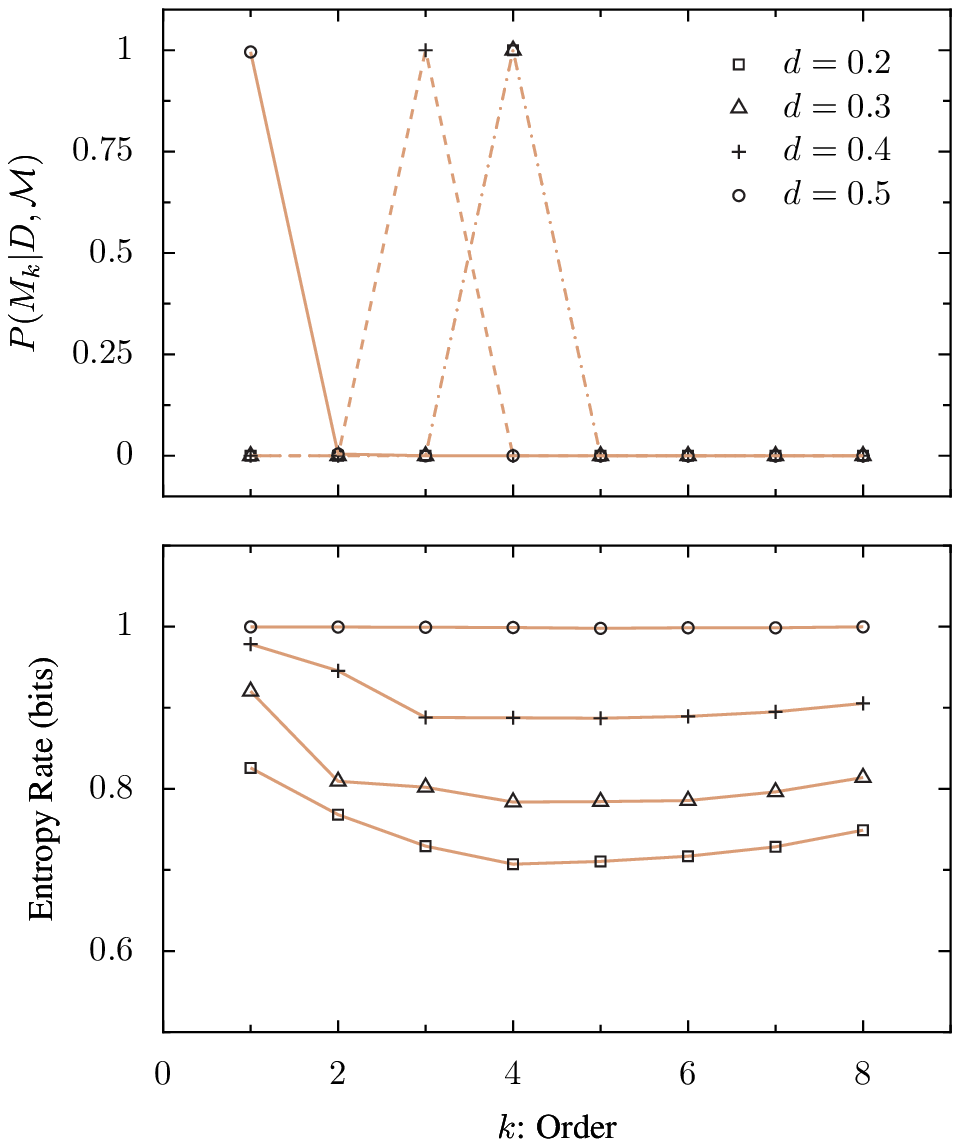}
}
\caption{Analysis of a single data stream of length $N=10^4$ from the
logistic map at $r=4.0$ with a noise level $\sigma = 10^{-3}$. Two hundred
evenly spaced decision points $d \in [0,1]$ were used to define measurement
partitions.}
\label{fig:sub} 
\end{figure}

The results of our experiments are presented in~\figref{fig:sub}. The bottom
panel of~\figref{fig:sub:a} shows the entropy rate $h_{\mu}(d)$ versus
decision point estimated using~\eqnref{eqn:average_info}. Note the nontrivial
$d$ dependence of $h_{\mu}(d)$. The dashed line shows an accurate numerical
estimate of the Lyapunov exponent using~\eqnref{eqn:lyapunov_exponent}.
It is also known to be $\lambda = 1$ bit per symbol from analytic results.
We note that $h_{\mu}(d)$ is zero at the extremes of $d=0$ and $d = 1$;
the data stream there is all $1$s or all $0$s, respectively.
The entropy rate estimate reaches a maximum at $d=1/2$.  For this decision
point the estimated entropy rate is approximately equal to the Lyapunov
exponent, indicating this instrument results in a generating partition and
satisfies Piesin's identity. In fact, this value of $d$ is also known to
produce a Markov partition.

The top panel of~\figref{fig:sub:a} shows the Markov chain order $k$ used to
produce the entropy rate estimate for each value of $d$. This dependence on
$d$ is also complicated in ways one might not expect. The order $k$ has two
minima (ignoring $d=0$ and $ d = 1$) at $d=1/2$ and $d = f^{-1}(1/2)$. These
indicate that the model size is minimized for those instruments. This is
another indication of the Markov partition for $r=4.0$ and $d=1/2$. These
results confirm that the maximum entropy-rate instrument produces the most
effect instrument for analysis of deterministic chaos in the presence of
dynamical noise. The model order is minimized at the generating partition.

Now let's consider the model order estimation process directly.
The bottom panel of~\figref{fig:sub:b} shows the estimated entropy rate
$h_{\mu}(k)$ versus model order for four different decision points. A relative
minimum in the entropy rate for a given $d$ selects the model order. This
reflects an optimization for the most structure and smallest Markov chain
representation of the data produced by a given instrument. The top panel in
this figure shows the model probability versus $k$ for the same set of
decision points, illustrating exactly this point. The prior over model orders,
which penalizes for model size, selects the Markov chain with lowest $k$
and smallest entropy rate.

\section{Conclusion}

We analyzed the degree of randomness generated by deterministic chaotic systems
with a small amount of additive noise. Appealing to the well developed theory
of symbolic dynamics, we demonstrated that this required a two-step procedure:
first, the careful design of a measuring instrument and, second, effective
model-order inference from the resulting data stream. The instrument should be
designed to be maximally informative and the model inference should produce
the most compact description in the model class. In carrying these steps out
an apparent conflict appeared: in the first step of instrument design, the
entropy rate was maximized; in the second, it was minimized. Moreover, it was
seen that instrument design must precede model inference. In fact, performing
the steps in the reverse order leads to nonsensical results, such as using
the one or the other extreme decision point $d = 0$ or $d = 1$.

The lessons learned are very simply summarized: Use all of the data and
nothing but the data. For deterministic chaos careful decision point analysis
coupled with Bayesian inference and model comparison accomplishes both of
theses goals.

\subsubsection*{Acknowledgments}

This research was supported by the Santa Fe Institute and the Computational Science and Engineering Center at the University of California at Davis.

\subsubsection*{References}

\renewcommand\refname{} 
\vspace{-0.5cm} 
\bibliographystyle{unsrt}
\bibliography{hrct}

\begin{thebibliography}{10}

\bibitem{Bollt2000}
E.~M. Bollt, T.~Stanford, Y.-C. Lai, and K.~Zyczkowski.
\newblock Validity of threshold-crossing analysis of symbolic dynamics from
  chaotic time series.
\newblock {\em Phys. Rev. Lett.}, 85(16):3524 -- 3527, 2000.

\bibitem{BLHao1998}
B.-L. Hao and W.-M. Zheng.
\newblock {\em Applied Symbolic Dynamics and Chaos}.
\newblock World Scientific, 1998.

\bibitem{Daw2002}
C.~S. Daw, C.~E.~A. Finney, and E.~R. Tracy.
\newblock A review of symbolic analysis of experimental data.
\newblock {\em Rev. Sci. Instr.}, 74(2):915 -- 930, 2003.

\bibitem{Crutchfield1982}
J.~P. Crutchfield and N.~H. Packard.
\newblock Symbolic dynamics of one-dimensional maps: Entropies, finite
  precision, and noise.
\newblock {\em Int. J. Theo. Phys.}, 21:433--466, 1982.

\bibitem{Crutchfield1983}
J~P. Crutchfield and N.~H. Packard.
\newblock Symbolic dynamics of noisy chaos.
\newblock {\em Physica D}, 7D:201--223, 1983.

\bibitem{Kolmogorov1958}
A.~N. Kolmogorov.
\newblock A new metric invariant of transitive dynamical systems and of
  endomorphisms of lebesgue spaces.
\newblock {\em Dokl. Akad. Nauk SSSR}, 119(5):861--864, 1958.

\bibitem{Kolmogorov1959}
A.~N. Kolmogorov.
\newblock On the entropy as a metric invariant of automorphisms.
\newblock {\em Dokl. Akad. Nauk SSSR}, 124(4):754--755, 1959.

\bibitem{Piesin1977}
Ya.~B. Piesin.
\newblock {\em Uspek. Math. Nauk.}, 32:55, 1977.

\bibitem{Baldi2001}
P.~Baldi and S.~Brunak.
\newblock {\em Bioinformatics: The Machine Learning Approach}.
\newblock MIT Press, 2001.

\bibitem{MacKay2003}
D.~J.~C. MacKay.
\newblock {\em Information Theory, Inference, and Learning Algorithms}.
\newblock Cambridge University Press, 2003.

\bibitem{Samengo2002}
I.~Samengo.
\newblock Estimating probabilities from experimental frequencies.
\newblock {\em Phys. Rev. E}, 65:046124, 2002.

\bibitem{Cover1991}
T.~M. Cover and J.~A. Thomas.
\newblock {\em Elements of Information Theory}.
\newblock Wiley-Interscience, 1991.

\end{thebibliography}

\end{document}